# An Evaluation Benchmark for Adverse Drug Event Prediction from Clinical Trial Results

Anthony Yazdani[1,*], Alban Bornet[1], Philipp Khlebnikov[2], Boya Zhang[1], Hossein Rouhizadeh[1], Poorya Amini[2] and Douglas Teodoro[1,*]

[1]Department of Radiology and Medical Informatics, Faculty of Medicine, University of Geneva, Geneva, Switzerland
[2]Risklick AG, Bern, Switzerland
[*]corresponding author(s): anthony.yazdani@unige.ch, douglas.teodoro@unige.ch

## Abstract

Adverse drug events (ADEs) are a major safety issue in clinical trials. Thus, predicting ADEs is key to developing safer medications and enhancing patient outcomes. To support this effort, we introduce CT-ADE, a dataset for multilabel ADE prediction in monopharmacy treatments. CT-ADE encompasses 2,497 drugs and 168,984 drug-ADE pairs from clinical trial results, annotated using the MedDRA ontology. Unlike existing resources, CT-ADE integrates treatment and target population data, enabling comparative analyses under varying conditions, such as dosage, administration route, and demographics. In addition, CT-ADE systematically collects all ADEs in the study population, including positive and negative cases. To provide a baseline for ADE prediction performance using the CT-ADE dataset, we conducted analyses using large language models (LLMs). The best LLM achieved an F1-score of 56%, with models incorporating treatment and patient information outperforming by 21%–38% those relying solely on the chemical structure. These findings underscore the importance of contextual information in ADE prediction and establish CT-ADE as a robust resource for safety risk assessment in pharmaceutical research and development.

## 1. Background & Summary

The development of pharmaceuticals faces numerous challenges, particularly the high incidence of adverse drug events (ADEs), which significantly contribute to the discontinuation of drug candidates[1]. ADEs are injuries resulting from medical intervention related to a drug, including those caused by the drug's pharmacological properties, improper dosage, or interactions with other medications, whether from appropriate use or misuse[2]. Data show that about 96% of drug candidates do not receive market approval, underscoring the inefficiencies and financial risks in drug development[3]. The average investment to bring a new drug to market is estimated at $1.3 billion, with costs for specific drugs varying widely depending on the therapeutic area[4]. A recent analysis shows that safety concerns are responsible for 17% of clinical trial (CT) failures[1], underscoring the critical need for improved predictive methods for managing ADEs. Such failures not only present substantial financial risks to pharmaceutical companies but also

raise ethical issues, especially considering the human costs associated with ADEs during CTs[3,5]. Drug candidates deemed safe in preclinical stages can exhibit toxic effects in clinical phases, leading to their failure. A notable factor contributing to this problem is the discrepancy between animal models used in preclinical screenings and human physiological reactions, indicating a significant gap in translating preclinical safety data to human contexts, which can result in severe ADEs, including fatalities[3,5–7]. In this context, in-silico models emerge as a promising approach for a safer and more accurate prediction of ADEs, potentially minimizing the differences observed between preclinical and clinical outcomes in pharmaceutical research and development.

Recent advancements in artificial intelligence and machine learning have drawn interest in this area, with research now focused on these technologies to complement existing methods in forecasting ADEs[8–19]. Early research efforts were centered on particular use cases, such as specific medications[8–11] and organ systems or routes of administration[12–14]. These methods have provided good explainability but have a limited range of applicability. To overcome these limitations, machine learning models that consider the molecular structure of drugs have been proposed[15–17]. These models work with the chemical space of drugs and are meant to enable predictions across a larger and more diverse set of compounds[20]. Drugs are encoded in standard representations such as SMILES[21], SELFIES[22], and molecular descriptors[23], and are associated with ADEs, such as those reported in public registries. Despite their sophistication, they often struggle to significantly outperform simpler approaches.

Existing benchmark datasets such as SIDER[24], AEOLUS[25], and OFFSIDES[26] have been used to analyze and predict drug-ADE associations using data-driven approaches. SIDER is a dataset comprising 1,430 unique drugs that compile ADEs reported in public documents and package inserts. It is designed through automated text mining and manual curation to link drugs with their reported ADEs. AEOLUS comprises 4,245 unique drugs and is derived from the FDA's adverse event reporting system (FAERS) (https://www.fda.gov/), standardizing ADE reports to facilitate analysis. This dataset focuses on post-marketing surveillance, offering a broad view of ADEs collected in real-world settings. OFFSIDES, a dataset composed of 1,332 unique drugs, identifies overlooked ADEs by analyzing data from FAERS, focusing on ADEs not listed on the official drug labels. Despite their significant contributions, these datasets are limited to approved treatment regimens and lack information from controlled environments. Specifically, they do not always account for the total number of patients treated, the precise proportion of those who experienced ADEs, or detailed patient characteristics and treatment regimens, altogether. Furthermore, no comparative cases exist where identical drugs are used under different conditions. Still, it is known that various contextual factors such as demographics, medical history, drug dosage, body weight, alcohol consumption, ethnicity, smoking habits, and pre-existing conditions influence the occurrence of ADEs[27].

To address these limitations, we developed CT-ADE[28], a comprehensive dataset that uniquely integrates five features not collectively available in existing resources: i) *Patient data*, encompassing information such as demographics, pathologies, and allergies, enabling the study of population-specific ADE risks; ii) *Treatment regimen data*, detailing information such as dosage, route, duration, and frequency of administration to improve regimen-specific predictions; iii) *Complete enumeration (census)* of ADE outcomes, systematically capturing all positive and negative cases within the study population, unlike voluntary reporting systems; iv) *Controlled monotherapy data*, derived from clinician-controlled trials that ensure strict adherence to treatment regimens while eliminating the confounding effects of polypharmacy; and v) *Comparative analysis opportunities*, allowing the study of identical drugs under varying conditions, such as patient demographics or treatment regimens. To the best of our knowledge, and as highlighted in a recent review[29], CT-ADE[28] is the first benchmark dataset to consider patient, drug, and treatment regimen data collectively.

CT-ADE[28] was compiled from CT results available through ClinicalTrials.gov (https://clinicaltrials.gov/), offering a rich resource for advancing risk assessment in pharmaceutical research and development. The dataset is structured to support a classification task, focusing on analyzing study groups within CTs that adhere to monopharmacy, i.e., the practice of using a single drug for treatment. In the dataset, study groups describing interventions and their respective regimens are enriched with molecular structure information of the drugs being used, linked via DrugBank[30], PubChem[31], and ChEMBL[32]. This approach enables a clearer understanding of how individual drugs and regimens can lead to patient-specific ADEs, free from the confounding effects of multiple concurrent medications and lack of census data. CT-ADE[28] is designed as a multilabel classification dataset to reflect that a single drug can cause multiple ADEs. This is achieved by standardizing clinician-reported ADEs from clinical trials, aligning them with the system organ class (SOC) and preferred term (PT) levels of the Medical Dictionary for Regulatory Activities (MedDRA) (https://www.meddra.org/). The dataset encompasses up to 2,497 unique drugs and 168,984 drug-ADE pairs, providing an extensive resource for predictive modeling. CT-ADE[28] comprehensively covers all system organ classes and drug pharmacological groups, offering a robust foundation for ADE prediction and enabling its application across diverse therapeutic areas and drug classes.

## 2. Methods

This section discusses the methodological framework for dataset creation, including the rationale for dataset splitting and quality assessment approaches. Moreover, it covers the selection and consolidation of source materials, data acquisition from CT results, DrugBank, PubChem, ChEMBL, and MedDRA ontology, and pre-processing steps for standardization.

### 2.1 CT-ADE resources

CT-ADE[28] dataset is based on five primary resources: ClinicalTrials.gov, DrugBank, PubChem, ChEMBL, and the MedDRA ontology.

**ClinicalTrials.gov**: ClinicalTrials.gov is a comprehensive registry of CTs maintained by the U.S. National Library of Medicine that provides up-to-date information on ongoing, completed, and terminated trials across a diverse range of drugs, diseases, and medical conditions. It offers transparency and access to detailed information on study objectives, design, methodology, eligibility criteria, locations, and sponsors. It describes treatment regimens, including the duration, strength, form, and dosage of interventions for each study group and their corresponding ADEs.

**DrugBank, PubChem, and ChEMBL:** These knowledgebases cover a wide range of compounds and drug properties, including their chemical structure information. DrugBank is maintained by the University of Alberta and The Metabolomics Innovation Center, PubChem by the National Center for Biotechnology Information, and ChEMBL by the European Bioinformatics Institute.

**MedDRA:** MedDRA is an internationally recognized medical terminology system used extensively by health authorities and the biopharmaceutical industry. It supports the standardized classification of adverse event data through a hierarchical system ranging from specific symptoms to broad organ system categories.

## 2.2 Data acquisition and pre-processing

In our pre-processing pipeline, we selected the following information from the data sources:

**ClinicalTrials.gov**: We selected CTs with completed or terminated status, involving at least one monopharmacy intervention, and with results reporting adverse events. The data were downloaded on April 17, 2024.

**DrugBank, PubChem, and ChEMBL**: We downloaded the DrugBank database version 5.1 on March 14, 2024, and two specific subsets from ChEMBL – Approved and USAN – on April 18, 2024. For PubChem, we specifically selected entries annotated as linked to ClinicalTrials.gov, ensuring relevance and integration with CT data. The PubChem data were downloaded on April 18, 2024. ATC codes and relevant synonyms were extracted from each database.

**MedDRA**: We selected MedDRA's English version 25.0. The MedDRA ontology was structured as a graph based on the hierarchical relationships from broader classifications (SOC) to detailed descriptions (LLT).

## 2.3 CT-ADE construction

As illustrated in Table 1, CT-ADE[28] provides a detailed representation of individual study group instances from CTs, including information about the i) *intervention name*, which indicates the name of the drug under investigation, ii) *ATC* codes, which classify drugs based on their areas of action, iii) *SMILES* notation, which provides a computational representation of the drug's chemical structure, iv) *eligibility criteria*, which specify the demographic and medical characteristics required for participant inclusion, thereby defining the target population for the intervention, v) *group description*, which describes the treatment regimen, including dosage and administration details, and vi) *ADE* label, which lists the ADE events associated with the groups at the SOC and PT MedDRA levels. Additionally, CT-ADE[28] includes CT-level information such as participant health statuses, gender, age group, and trial phase. The dataset is segmented into two versions, i.e., CT-ADE-SOC[28] and CT-ADE-PT[28]. All versions share the same features, but ADE labels vary depending on the MedDRA target level, indicating the occurrence (1) or absence (0) of statistically significant ADEs.

| Intervention Name | ATC | SMILES | Eligibility Criteria | Group Description | ADE |
|---|---|---|---|---|---|
| Tasimelteon | N05CH03 | CCC(=O)NC[C@@H]1C[C@H]1c1cccc2c1CCO2 | […] Males and females with a diagnosis of primary insomnia […] | 20 mg VEC-162 capsules, PO daily for five weeks | Infections and infestations<br><br>Nervous system disorders |

Table 1. Features available for a single instance in the CT-ADE[28] dataset. Some features are not shown to improve readability. Omitted features include participant health status, gender, age group, participant count, and trial phase.

### 2.3.1 Deconstructing clinical trials

The unique configuration of CT-ADE[28] enables a single CT to generate multiple data entries. As CTs can evaluate multiple drugs or distinct treatment regimens, separate data entries for each study group are required to capture this information. Study groups are defined by their specific intervention strategies and help study how variations in drugs or regimen details – such as dosage, administration, and duration – contribute to differing ADE profiles for the same population (i.e., same eligibility criteria). We developed a preprocessing pipeline to systematically deconstruct CTs into study groups, ensuring the accurate representation of group-specific ADE data. The pipeline processes CTs sourced from ClinicalTrials.gov, focusing exclusively on monopharmacy interventions and filtering for trials with completed or terminated statuses, classified as interventional, and reporting results.

As illustrated in Figure 1, which depicts the CT preprocessing pipeline, the protocol section of each CT outlines the eligibility criteria shared across study groups, defining the target population. The trial is then divided into study groups based on distinct intervention strategies, each represented by a triplet ("arm group title," "arm group description," "raw intervention name"). The result section complements this by describing ADE data for each group, forming a second triplet ("ADE group title," "ADE group description," "ADE group report").

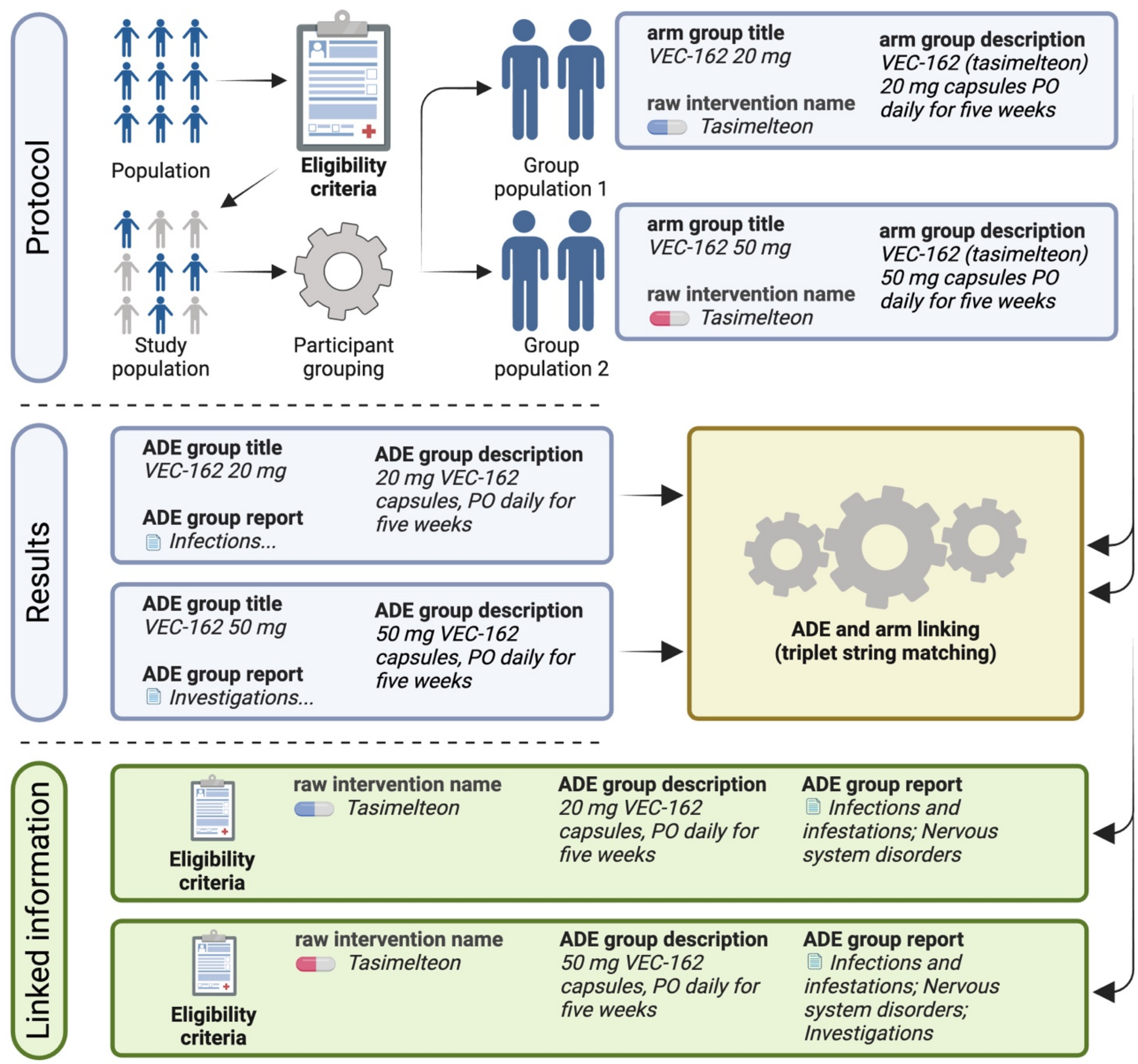

Figure 1. Linking protocol and result sections of clinical trials to generate raw CT-ADE[28] instances. The protocol section provides eligibility criteria and intervention details, while the result section reports ADE outcomes. The pipeline matches these sections to create structured associations between interventions and their ADEs.

The primary goal of the deconstruction process is to link protocol and result triplets into a dataset that combines CT-level metadata, intervention details, and group-specific ADE outcomes. Since direct links between intervention details and ADE reports are absent in raw data, the pipeline employs string matching between protocol triplets ("arm group title," "arm group description," "raw intervention name") and result triplets ("ADE group title," "ADE group description," "ADE group report"). This ensures accurate associations between interventions and ADEs.

In single-intervention CTs, linking protocol and result triplets is straightforward. The pipeline matches the raw intervention name with the ADE group title or description using inclusion matching. Matches are accepted only when the intervention name uniquely appears in one ADE group, ensuring unambiguous associations. For trials with multiple interventions, the pipeline first identifies unique matches between arm and ADE group triplets. If an arm group title or description corresponds to only one ADE group title or description, the two are linked. When multiple matches occur, stricter criteria are applied – both the arm group title and description must align with a single ADE group.

The pipeline excludes instances that fail to meet unique match criteria, ensuring the final dataset contains only strict matches. Each successfully matched study group inherits CT-level metadata, such as eligibility criteria, participant health status, gender, age group, and trial phase, further enriching the dataset with comprehensive contextual information. This approach reliably captures multiple- and single-intervention CTs while maintaining high data integrity. By implementing strict matching criteria and systematically linking protocol and result triplets, the CT preprocessing pipeline provides a raw dataset that reflects the nuanced effects of patient- and regimen-level information on ADEs. Among the 491,535 clinical trials available on ClinicalTrials.gov as of April 17, 2024, 61,921 met the criteria of being completed or terminated, interventional, and reporting results. Of these, we extracted and linked ADE data on 31,419 monopharmacy study groups.

#### 2.3.2 Standardizing raw intervention names

Deconstructing CTs provides essential information, including raw intervention names, eligibility criteria, group descriptions, participant health status, gender, age group, trial phase, and raw ADEs. However, to achieve fully mapped instances, as illustrated in Table 1, it is crucial to convert the raw intervention names into standardized representations. This process involves mapping the raw intervention names to their respective canonical names, ATC codes, and SMILES. Initially, we consolidated information from DrugBank, PubChem, and ChEMBL, unifying these knowledge bases by grouping identical compounds by their canonical SMILES or names, and merging their synonyms. Then, we performed an exact and inclusion match of raw intervention names against this unified database. If no match was found for a given instance, we normalized the raw intervention name by removing dosage details, forms, and routes of administration, and by cleaning the text from special characters to enhance compatibility (e.g., transforming "Diprosone® Cream 0.05%" into "diprosone"). Then, we repeated the matching process. From the 31,419 monopharmacy study groups obtained through CT deconstruction, encompassing 13,110 unique raw intervention names, 7,081 were successfully mapped to their standardized representations. This mapping resulted in 2,825 unique drugs and 21,306 study groups with mapped interventions, populating the Intervention Name, ATC, and SMILES columns of Table 1.

#### 2.3.3 Standardizing adverse drug events

The MedDRA ontology provides a comprehensive framework for standardizing ADE concepts, enabling reliable comparison and aggregation of ADE data across studies. Each of the 21,306 intervention-mapped study groups provides a census of ADE events, i.e., each reported ADE includes its term, the affected organ system, the number of affected patients, and the total number of participants in that study group. To harmonize this data, we standardized the reported ADEs by mapping each ADE term to its corresponding PT concept and linking each organ system to its respective SOC concept. This standardization process employed strict string matching to maintain precision and consistency.

Building on this standardized dataset, we ensured that ADE labels were statistically significant and clinically relevant. To achieve this, we used the Wilson interval for binomial proportion confidence[33]. The Wilson method ensures asymmetric confidence intervals and constrains the boundaries to a valid probability range. It is particularly suitable for near-boundary estimation and small population studies[34], as is common in the early phases of clinical trials[35]. However, the Wilson lower bound never reaches zero when there is at least one observed event, posing a challenge for clinical datasets, where any occurrence – no matter how rare – is always flagged as non-zero. To address this, we labeled an ADE occurrence as positive if at least 1% of the population was affected with 95% confidence. We applied this threshold following internationally recognized standards from the Council for International Organizations of Medical Sciences[36]. These guidelines define common ADEs as those occurring in at least 1% of the population, ensuring that positive labels are both statistically significant and clinically relevant. Study groups with any unmapped positive ADEs were excluded to prevent the assignment of false negatives. However, groups reporting no ADEs, i.e., indicating that no ADEs have occurred, were retained.

Among the 21,306 intervention-mapped study groups, 103 were excluded due to missing data on the number of affected patients or total participants, preventing the application of the Wilson interval for statistical evaluation. Additionally, 6,006 study groups at the SOC level and 5,563 at the PT level were excluded due to the absence of strict matches to MedDRA concepts.

### 3. Data Records

CT-ADE[28] is available on Figshare (https://figshare.com/articles/dataset/28142453) and HuggingFace (https://huggingface.co/anthonyyazdaniml). The dataset is organized into two distinct versions – CT-ADE-SOC[28] and CT-ADE-PT[28]. Both versions are divided into training, validation, and test sets, stratified to ensure no common drugs are shared between splits, thereby avoiding data leakage. This stratification is consistent across levels, ensuring that drugs in the SOC training split are not found in the validation or test splits of the PT version.

The dataset is organized into the following directory structure:

- ct_ade/soc/: Contains files for the SOC-level dataset.
    - train.csv: SOC-level training data.
    - val.csv: SOC-level validation data.
    - test.csv: SOC-level test data.

- train_frequencies.csv: Raw ADE frequency data for the training split.
- val_frequencies.csv: Raw ADE frequency data for the validation split.
- test_frequencies.csv: Raw ADE frequency data for the test split.

- ct_ade/pt/: Contains files for the PT-level dataset, with a structure identical to the SOC-level directory.

Each split file (train.csv, val.csv, and test.csv) includes metadata about clinical trials, drug information, and ADE labels:

- nctid: Identifier for the clinical trial.
- group_id: Identifier for study groups within a trial.
- healthy_volunteers: Indicates whether the study involves healthy volunteers.
- gender: Participant gender category.
- age: Participant age category.
- phase: Clinical trial phase.
- ade_num_at_risk: Number of participants in the study group.
- eligibility_criteria: Eligibility criteria for participant selection.
- group_description: Description of the treatment regimen.
- drug_info_source: Link of the drug under study to DrugBank, PubChem, and/or ChEMBL.
- intervention_name: Name of the drug under study.
- smiles: SMILES representation of the drug's chemical structure.
- atc_code: ATC classification code for the drug.
- label_*: Binary indicators (1 for presence, 0 for absence) of ADEs, according to MedDRA SOC or PT levels.

The frequency files (train_frequencies.csv, val_frequencies.csv, and test_frequencies.csv) provide quantitative details about the occurrence of ADEs:

- The frequency files include the nctid and group_id columns to enable linkage with the split files (train.csv, val.csv, and test.csv) but replace label_* columns with frequency_* columns.
- Each frequency_* column provides raw ADE frequency data, reflecting the proportion of participants experiencing a specific ADE. Frequencies are omitted if the association between an ADE concept and its frequency cannot be confidently established.

Table 2 shows the statistics of the CT-ADE[28] dataset across SOC and PT levels for the training, validation, and test splits. The number of drug-ADE pairs in the dataset varies between 40,187 at the SOC level and 168,984 at the PT level, including a maximum of 2,497 unique drugs associated with 15,640 study groups at the PT level.

| Level | Split | Unique drugs | Study groups | Drug-ADE pairs |
|---|---|---|---|---|
| SOC | Train | 1,992 | 12,419 | 32,191 |
| | Validation | 244 | 1,518 | 4,146 |
| | Test | 238 | 1,260 | 3,850 |
| | **Total** | **2,474** | **15,197** | **40,187** |
| PT | Train | 2,000 | 12,736 | 132,917 |
| | Validation | 247 | 1,509 | 17,458 |
| | Test | 250 | 1,395 | 18,609 |
| | **Total** | **2,497** | **15,640** | **168,984** |

Table 2. Summary of key metrics from CT-ADE[28].

## 4. Technical Validation

### 4.1 Quality control

Accurate mapping of raw intervention names to SMILES representations available in drug knowledge bases (DrugBank, PubChem, and ChEMBL) was critical to guarantee quality. To minimize mapping errors, we only used strict matching techniques. Similarly, the mapping strategy for ADE terms reported in CTs to the MedDRA ontology was performed using an exact match. To further improve the dataset's integrity, we excluded study groups where we could not map all positive ADEs, preventing the inclusion of incomplete data. Moreover, only ADE reports documenting all necessary fields – specifically, the adverse event, the number of affected individuals, and the total patient count in the study group – were included. To ensure that the positive ADEs were statistically significant and clinically relevant, the Wilson interval for binomial proportion confidence was used to assign classification labels to ADEs. Specifically, an ADE occurrence was assigned a value of 1 if we were 95% confident that at least 1% of the population would experience the ADE; otherwise 0.

### 4.2 Dataset coverage

This section evaluates the extent and diversity of the CT-ADE-SOC[28] dataset by analyzing the distribution of ADEs across SOCs and ATC main pharmacological groups of the included drugs. As shown in Figure 2A, all the 27 SOC categories of MedDRA are covered in the CT-ADE-SOC[28] dataset, with the top-3 most represented SOCs being "Gastrointestinal disorders" (Gastr) (38.33%), "Nervous system disorders" (Nerv) (34.49%) and "Infections and infestations" (Infec) (26.85%), and the top-3 least represented being "Product issues" (Prod) (0.33%), "Congenital, familial and genetic disorders" (Cong) (0.28%) and "Social circumstances" (SocCi) (0.19%). Similarly, Figure 2B shows that all ATC main pharmacological groups are included in the dataset. The top-3 most represented groups are "Nervous System" (N) with 3,152 instances, "Alimentary Tract and Metabolism" (A) with 2,590 instances, and "Cardiovascular System" (C) with 1,486 instances. On the other hand, the top-3 least represented groups are “Systemic Hormonal Preparations, Excl. Sex Hormones and Insulins” (H) with 536 instances, “Various” (V) with 459 instances, and “Antiparasitic Products, Insecticides and Repellents” (P) with 154 instances. For brevity, we present the analysis based on the full CT-ADE-SOC[28] dataset, but this coverage is consistent across all SOC splits. A similar analysis for CT-ADE-PT[28] is shown in Supplementary Figure 1.

To validate the representativeness of the dataset, we compared the SOC-level frequencies of ADEs in CT-ADE-SOC[28] to global population frequencies reported in a large-scale study by Aagaard *et al.*[37]. Our analysis revealed a Spearman correlation coefficient of 0.862 (p-value < 0.001), indicating a strong and statistically significant alignment between CT-ADE[28] and global ADE patterns.

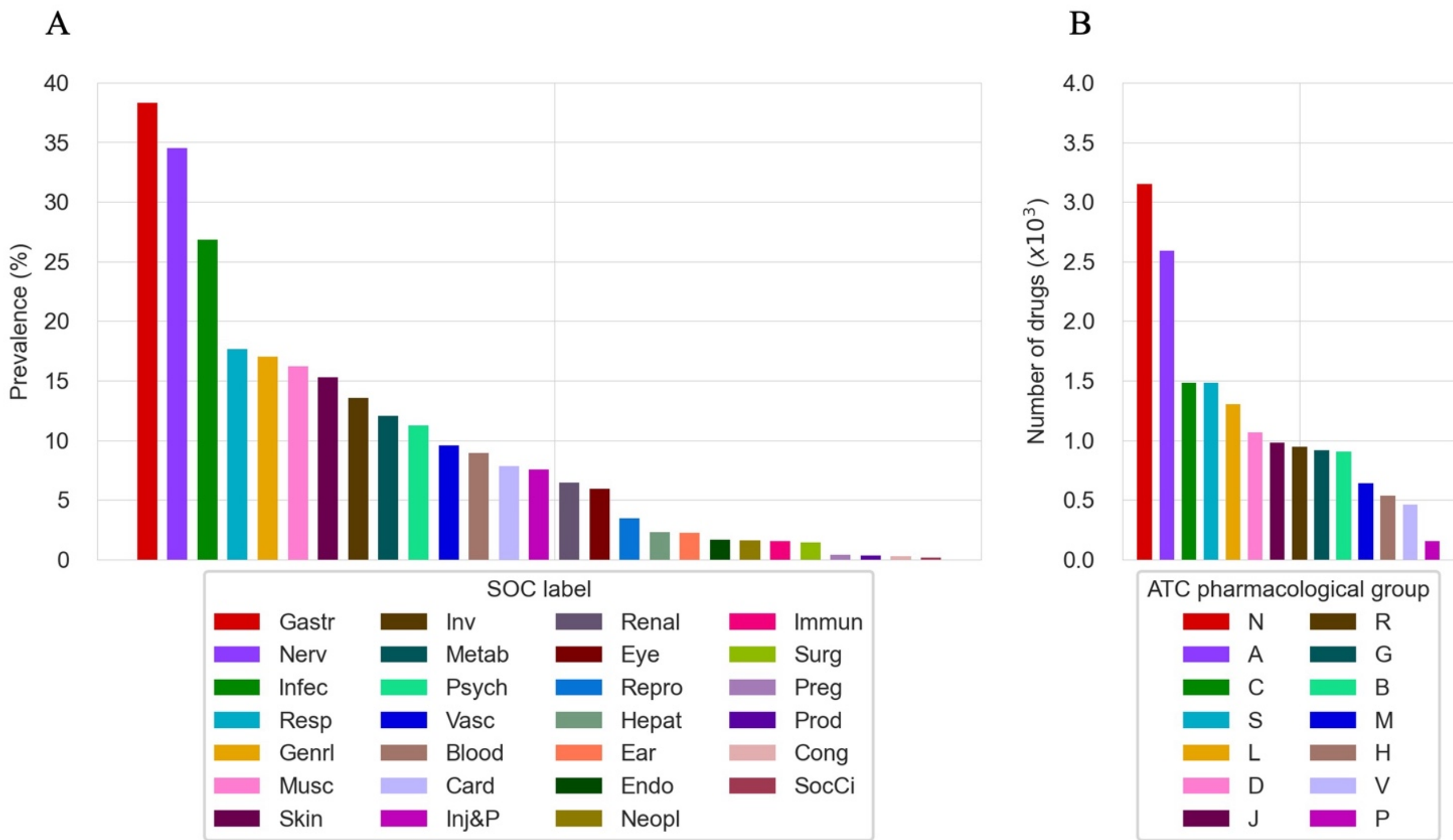

Figure 2. (A) Distribution of SOCs in CT-ADE-SOC[28]. (B) Representation of ATC main pharmacological groups in CT-ADE-SOC[28]. Abbreviation mappings for SOCs and ATC main pharmacological groups to their full terms are available in Supplementary Tables 4 and 5, respectively.

### 4.3 Experiments

To test the hypothesis that adding patient and treatment regimen information enhances ADE prediction, we conducted experiments with discriminative and generative large language models (LLMs). These experiments established baseline performance and evaluated the added value of contextual information compared to the standard use of chemical structure data. Since CT-ADE[28] emphasizes patient-specific and treatment-related data, we focused our technical validation on LLMs pre-trained on biomedical corpora. We systematically tested three feature configurations:

- SMILES only (S) configuration, focusing solely on the SMILES notation of drug compounds.
- SMILES and group description (SG) configuration, which incorporates group descriptions to exploit both chemical properties and treatment regimens.
- SMILES, group description, and eligibility criteria (SGE) configuration, providing a detailed context for ADE prediction by including the target population information.

To quantify the incremental improvements achieved by progressively adding contextual features, we employed the micro-averaged McNemar's test. This method evaluates the impact of each feature configuration (S vs. SG and SG vs. SGE) on predictive performance, highlighting the contribution of each additional layer of contextual information. Discriminative models used distinct encoders for each feature modality (Figure 3). The S configuration used ChemBERTa-77M-MLM[38] as a backbone to encode the SMILES notations. The SG configuration combined ChemBERTa-77M-MLM with PubMedBERT-base[39] backbones, integrating the treatment regimen alongside the SMILES notations. The SGE configuration provided the most complete set of input features by incorporating patient eligibility criteria and used the same backbones as for SG.

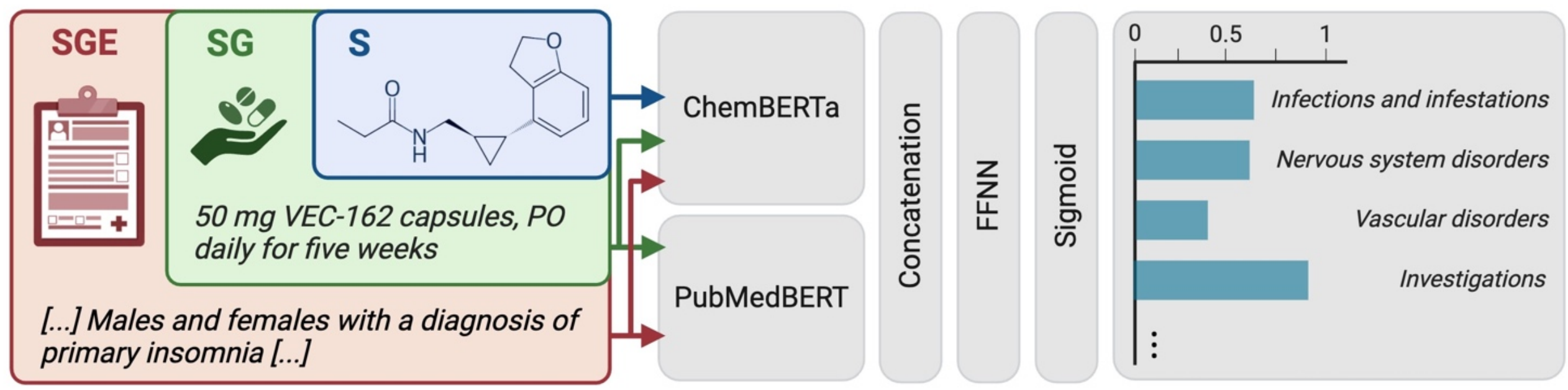

Figure 3. Discriminative model pipeline used for ADE prediction. The model uses dedicated encoders for text (handling both eligibility criteria and group descriptions sequentially) and SMILES strings. The encoded features are concatenated and sent to a feed-forward neural network. Final output probabilities are computed using the sigmoid activation function.

Generative models were based on OpenBioLLM-8B[40], an open-source biomedical instruction model, which was used with the official chat template. Similar to the discriminative approach, the models were fine-tuned to generate a list of ADEs based on S, SG, and SGE scenarios (Figure 4). OpenBioLLM-8B was fine-tuned with bf16 precision[41], using low-rank adapters[42], flash attention 2[43], and gradient checkpointing[44] in a completion-only framework.

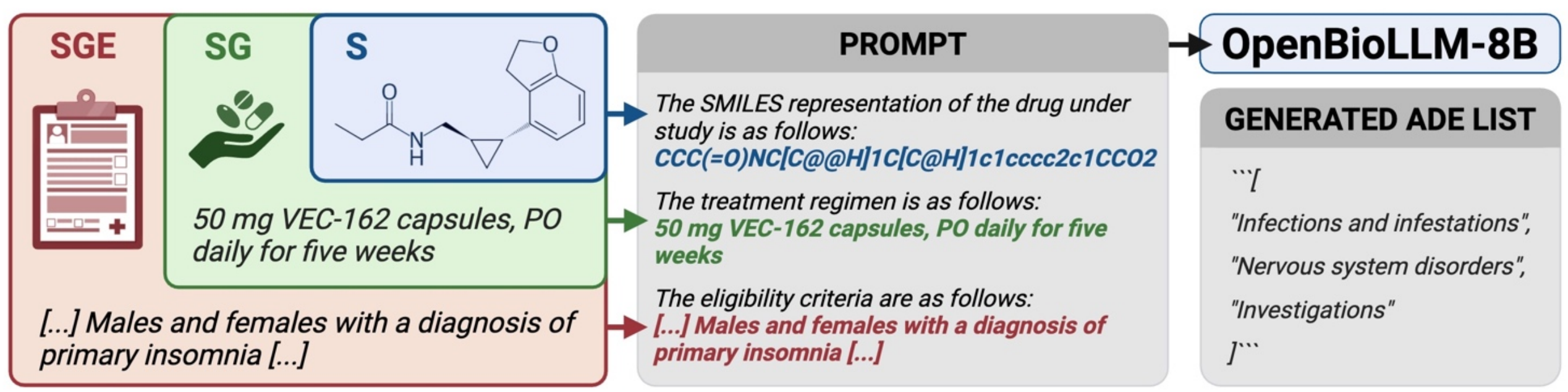

Figure 4. Generative model pipeline used for ADE prediction. A prompt is created based on the input features, and OpenBioLLM-8B is tasked to generate a list of ADEs.

As shown in Figure 5, for the S configuration, the discriminative model achieves an F1-score of 31.96%, while the generative model achieves an F1-score of 24.16%. The results establish a baseline performance, providing a reference point to assess the incremental contributions of patient- and treatment-specific information. For the SG configuration, the discriminative model improves significantly compared to the S configuration, with a micro F1-score of 46.09% (p-value < 0.001), whereas the generative model achieves a micro F1-score of 49.74% (p-value = 0.15). This demonstrates the substantial impact of integrating treatment regimen information, balancing precision and recall, and enhancing predictive performance. Finally, the SGE configuration provides the best performance for both model types. Compared to the SG configuration, the discriminative model achieves an F1-score of 53.46% (p-value = 0.11), while the generative model achieves an F1-score of 53.43% (p-value < 0.001).

Since only 11.32% of instances are positive in the CT-ADE-SOC[28] test set, we also tested the models on the subset where ADEs were observed, focusing on their ability to identify positive cases. In this scenario, the performance differences for any comparison – S vs. SG, S vs. SGE, and SG vs. SGE – are statistically significant (p-values < 0.001) for both discriminative and generative models. These results demonstrate the improvements in ADE prediction achieved by adding contextual information. This in-silico analysis is consistent with findings from existing in-vivo models[27] and highlights the critical role of incorporating patient- and treatment-specific information to improve ADE prediction. Similar conclusions hold for the CT-ADE-PT[28] dataset, with detailed results using discriminative models provided in Supplementary Tables 1 and 2.

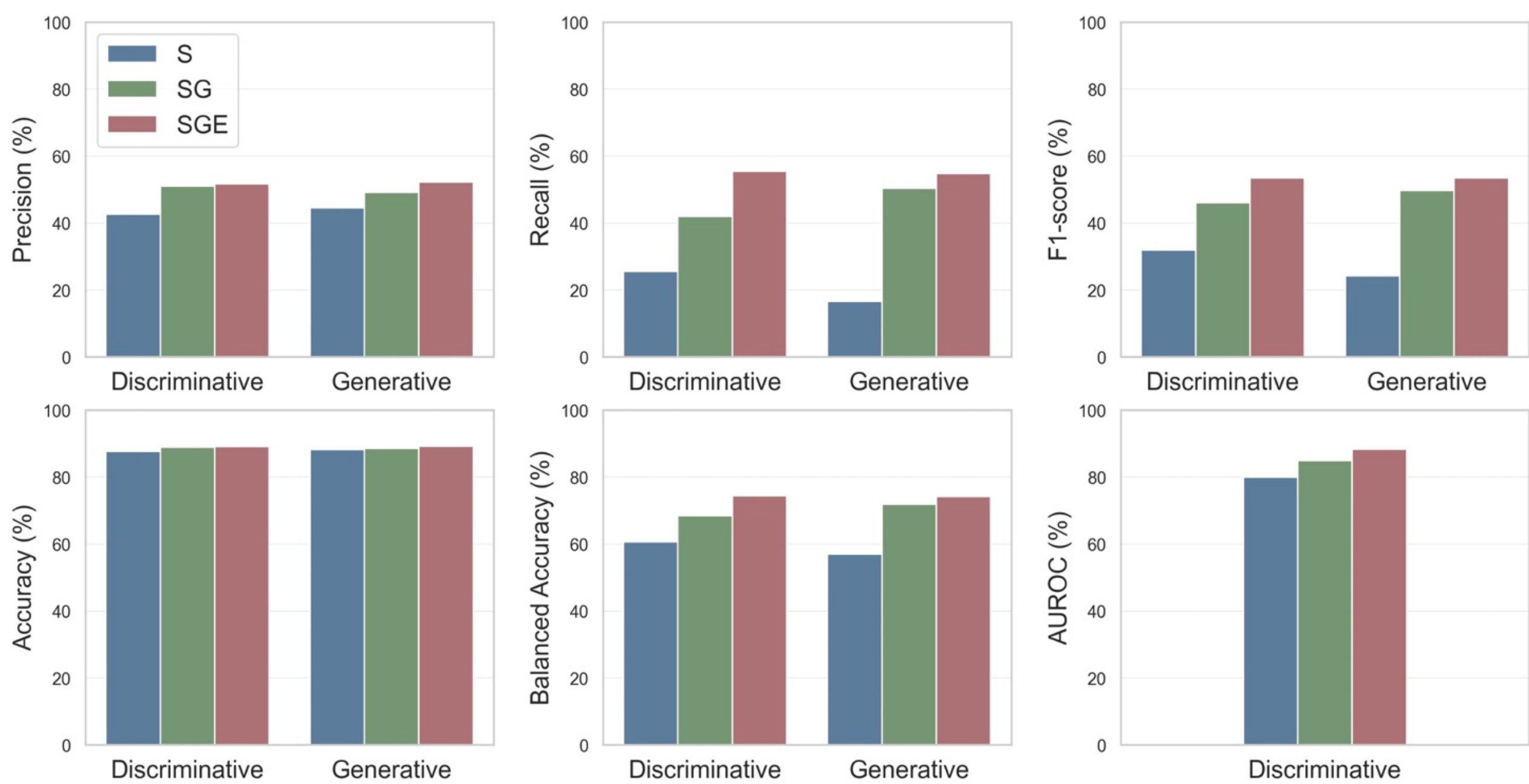

Figure 5. Performance comparison of discriminative (ChemBERTa-77M-MLM & PubMedBERT) and generative (OpenBioLLM-8B) models on the CT-ADE-SOC[28] test split using different feature sets (S, SG, SGE). The AUROC metric cannot be computed for the generative model as it does not produce raw probabilities. All metrics are micro-averaged. Tabular values are available in Supplementary Table 3.

### 4.3.1 Alternative evaluation scenarios

Performance results across the SOC levels and ATC main pharmacological groups for the best discriminative model (SGE) are provided in Figure 6. At the SOC level, the discriminative SGE model demonstrates strong performance in predicting common ADEs. For example, it achieves an F1-score of 71.95% for "Gastrointestinal disorders" (Gastr) and 71.28% for "Nervous system disorders" (Nerv). However, the model's performance is weaker for rarer SOCs, such as "Social circumstances" (SocCi), where it fails to predict any ADE in this category. Similarly, performance is weaker for "Cardiac disorders" (Card) with an F1-score of 31.50%, despite their relative frequency. This suggests that, while the model handles common ADEs, refinement is needed to improve minority class prediction performance. Performance across ATC main pharmacological groups further highlights the model's strengths and areas for improvement. The SGE model achieves higher F1-scores in categories like "Blood and Blood Forming Organs" (B) (55.17%) and "Systemic Hormonal Preparations" (H) (58.54%). Conversely, the model performs worse in predicting ADEs for "Antiinfectives for Systemic Use" (J), with an F1-score of 32.38%. These insights emphasize the challenge of predicting ADEs in clinical research, and the importance of novel strategies to address specific ADE categories and therapeutic areas beyond LLM fine-tuning.

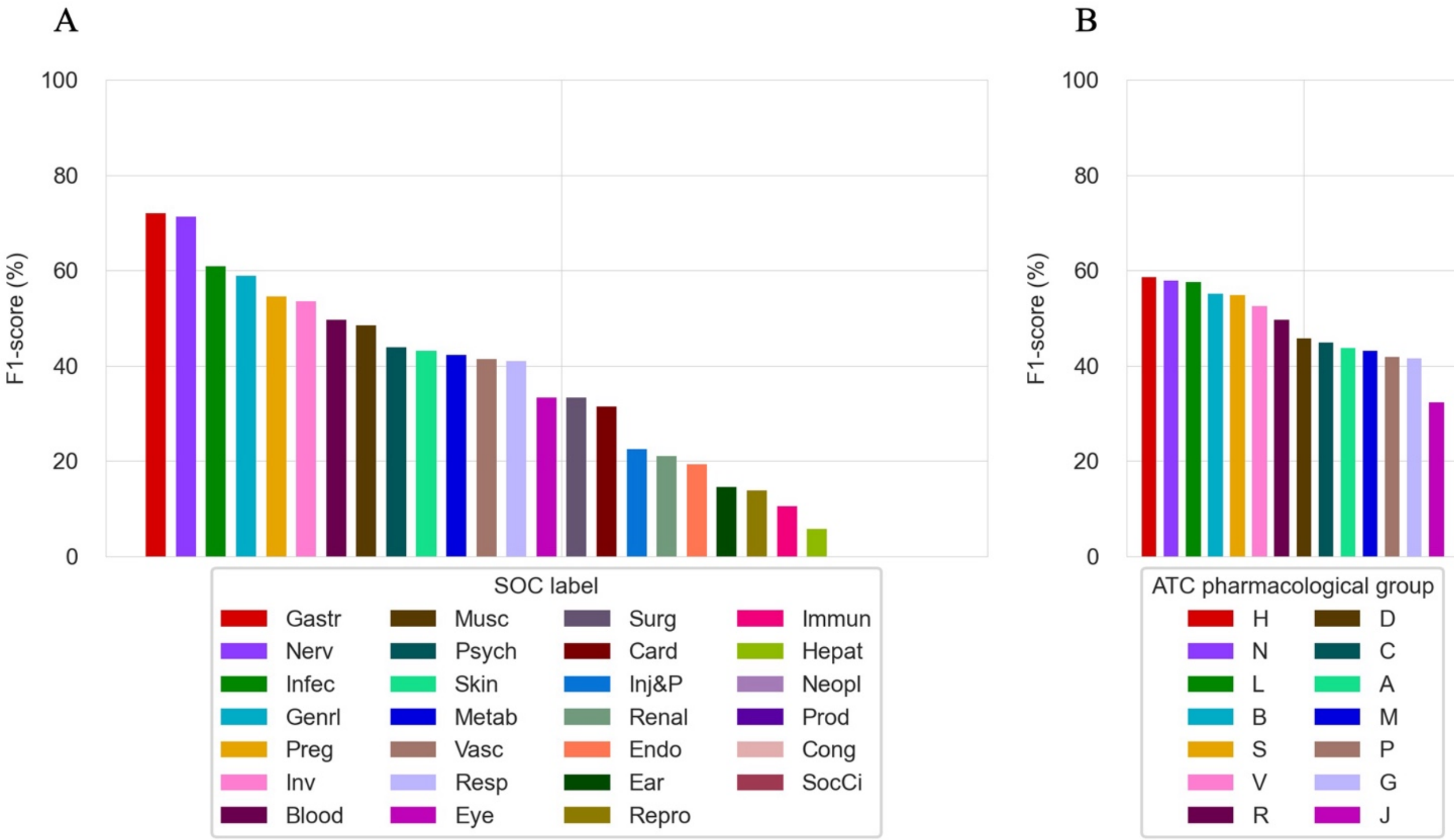

Figure 6. (A) F1-score of the SGE discriminative model on individual SOCs in CT-ADE-SOC[28] test set. (B) F1-score of the SGE discriminative model on individual ATC main pharmacological groups in CT-ADE-SOC[28] test set. Supplementary Tables 6 and 7 provide tabular values for the results shown in A and B and additional metrics. Abbreviation mappings for SOCs and ATC main pharmacological groups to their full terms are available in Supplementary Tables 4 and 5, respectively.

### 4.3.2 Effect of model scaling and domain pre-training on ADE prediction

To assess the impact of model scaling and domain-specific pre-training on ADE prediction, we additionally fine-tuned a range of generative models, including Llama-3[45] (8B, 70B), Meditron[46] (7B, 70B), and OpenBioLLM-70B. Due to computational constraints, these experiments were restricted to the full feature set (SGE). Performance was evaluated relative to our best discriminative model (SGE) and a baseline using a majority-class prediction approach (MAJ), which assumes no ADEs occur. As shown in Table 3, the Llama-3-8B model obtains the highest performance, with an F1-score of 55.90%, which is 2.4 percentage points above the SGE discriminative model (p-value $< 0.01$). Interestingly, despite their substantial parameter count – 70 times larger than the discriminative model – 8B generative models achieve comparable performance. This suggests that increasing the number of parameters does not necessarily lead to proportional performance improvements. As illustrated in Table 3, the 70B models demonstrate even more diminishing returns with parameter scaling. Although biomedical LLMs have been shown to outperform general domain models in biomedical tasks[47], we found that specialized domain models such as Meditron and OpenBioLLM do not provide a performance advantage compared to general domain models in CT-ADE-SOC[28]. Due to the imbalanced nature of ADE datasets, the majority class model (MAJ) tends to achieve strong performance in terms of accuracy. Llama-3-8B, with an accuracy of nearly 90%, improves only 0.89 percentage points upon the MAJ model. However, it can identify around 58% of ADEs, while the MAJ model does not predict any (p-value < 0.001). These findings suggest that increasing model sizes or pre-training models on domain-specific corpora does not necessarily improve ADE predictive performance for this task.

| **Model Type** | **Parameters (x10$^9$)** | **Backbone** | **Precision (%)** | **Recall (%)** | **F1-score (%)** | **Accuracy (%)** | **Balanced Accuracy (%)** |
|---|---|---|---|---|---|---|---|
| MAJ | 0 | - | 0.00 | 0.00 | 0.00 | 88.68 | 50.00 |
| Discriminative | 0.11 | ChemBERTa & PubMedBERT | 51.65 | 55.40 | 53.46 | 89.08 | 74.39 |
| Generative | 7 – 8 | Meditron | 52.82 | 53.84 | 53.32 | 89.33 | 73.85 |
| | | OpenBioLLM | 52.18 | 54.75 | 53.43 | 89.20 | 74.17 |
| | | Llama-3 | 53.60 | **58.42** | **55.90** | 89.57 | **75.98** |
| | 70 | Meditron | 61.01 | 44.10 | 51.20 | 90.49 | 70.25 |
| | | OpenBioLLM | 60.28 | 42.42 | 49.79 | 90.32 | 69.42 |
| | | Llama-3 | **62.09** | 49.30 | 54.96 | **90.86** | 72.73 |

Table 3. Performance metrics of various models using the SGE feature set evaluated on the CT-ADE-SOC[28] test split. All metrics are micro-averaged.

### 4.4 Limitations

Several limitations present opportunities for refinement of the CT-ADE[28] dataset. Firstly, CT-ADE[28] does not incorporate preclinical information, such as data from in vitro assays, into its drug features. However, by linking chemical databases like DrugBank, PubChem, and ChEMBL to study groups, CT-ADE[28] establishes a foundation for future integration of preclinical data, which may enhance the predictive power of ADE models. Secondly, since CT-ADE[28] is derived from controlled CT settings, it may not fully capture real-world variability due to strict inclusion and exclusion criteria, standardized treatment regimens, and closely monitored conditions. These characteristics, while ensuring data consistency and reliability, also limit the dataset's ability to reflect the complexities of routine medical practice. Additionally, CT-ADE[28] focuses exclusively on monopharmacy interventions, facilitating precise ADE attribution but excluding polypharmacy scenarios that are common in clinical practice. Expanding the dataset to encompass polypharmacy cases would enable models to account for drug-drug interactions and more complex treatment regimens, thereby enhancing their applicability to real-world settings. Lastly, the dataset is restricted to drugs with SMILES representations, thereby excluding compounds that lack such encodings due to their structural complexity. Incorporating alternative representations, such as amino acid sequences for biologics, could extend the applicability of CT-ADE[28] to a broader range of therapeutic agents.

## Code Availability

The code used to generate and evaluate the CT-ADE[28] dataset is publicly available on GitHub at https://github.com/ds4dh/CT-ADE. The repository includes all Python scripts and documentation required to reproduce the dataset and conduct the experiments described in this study. The code is released under the MIT license, and there are no further restrictions on its use.

## Author Contributions

D.T., P.A., and A.Y. conceptualized the study. A.Y. and A.B. implemented the codes for the creation and evaluation of the dataset. A.Y. and P.K. analyzed the results. The manuscript was drafted by A.Y. and edited by A.B. and D.T. All authors reviewed and approved the final version.

## Competing Interests

The authors declare the following competing interests: P.K., and P.A. work for Risklick AG. All other authors declare no competing interest.

# Supplementary information

**An Evaluation Benchmark for Adverse Drug Event Prediction from Clinical Trial Results**

Anthony Yazdani[1,*], Alban Bornet[1], Philipp Khlebnikov[2], Boya Zhang[1], Hossein Rouhizadeh[1], Poorya Amini[2] and Douglas Teodoro[1,*]

[1]Department of Radiology and Medical Informatics, Faculty of Medicine, University of Geneva, Geneva, Switzerland

[2]Risklick AG, Bern, Switzerland

[*]corresponding author(s): anthony.yazdani@unige.ch, douglas.teodoro@unige.ch

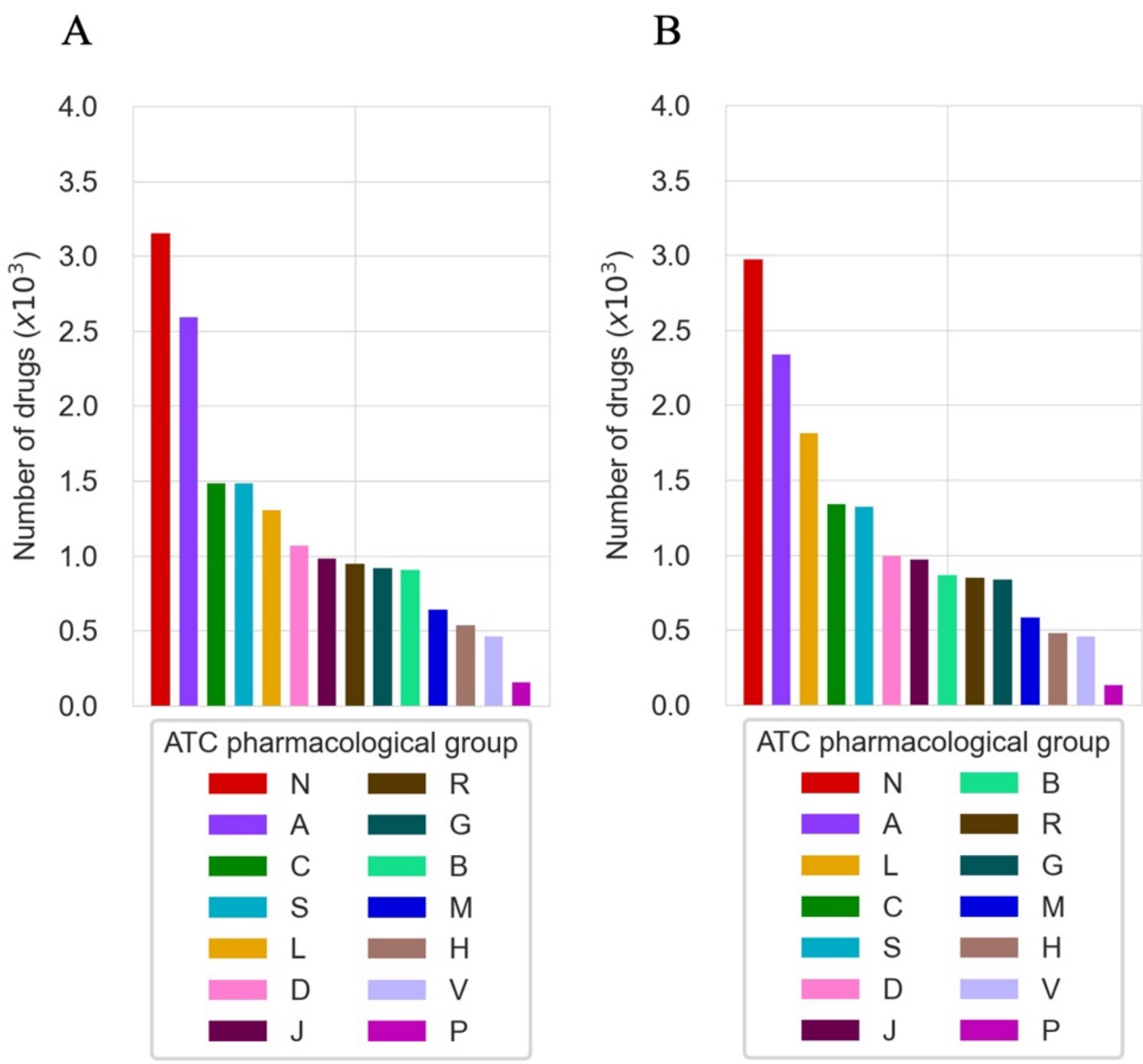

Supplementary Figure 1. (A) Representation of ATC main pharmacological groups in CT-ADE-SOC. (B) Representation of ATC main pharmacological groups in CT-ADE-PT.

| MedDRA Level | Model | Precision (%) | Recall (%) | F1-score (%) | Accuracy (%) | Balanced Accuracy (%) | AUROC (%) |
|---|---|---|---|---|---|---|---|
| PT | MAJ | 0.00 | 0.00 | 0.00 | **99.95** | 50.00 | - |
| | S | 0.09 | 2.78 | 0.17 | 98.32 | 50.57 | 73.18 |
| | SG | 48.59 | 24.50 | 32.58 | **99.95** | 62.25 | 98.04 |
| | SGE | **49.63** | **31.29** | **38.38** | **99.95** | **65.64** | **98.23** |

Supplementary Table 1. Performance of discriminative models on the CT-ADE-PT test set. MAJ: Majority class prediction; S: SMILES; SG: SMILES and group description; SGE: SMILES, group description, and eligibility criteria. Results are micro-averaged.

| MedDRA Level | Comparison | Full test set | Positive subset |
|---|---|---|---|
| PT | S vs. SG | < .001 | < .001 |
| | S vs. SGE | < .001 | < .001 |
| | SG vs. SGE | 0.05 | < .001 |

Supplementary Table 2. P-values from pairwise McNemar's tests comparing the performance of discriminative models on the CT-ADE-PT test set using different feature sets. The tests were conducted on the entire test set (Full test set) and the subset of the test set where ADEs were observed (Positive subset). S: SMILES; SG: SMILES and group description; SGE: SMILES, group description, and eligibility criteria.

| Type | Backbone | Features | Precision (%) | Recall (%) | F1-score (%) | Accuracy (%) | Balanced Accuracy (%) | AUROC (%) |
|---|---|---|---|---|---|---|---|---|
| MAJ | - | - | 0.00 | 0.00 | 0.00 | 88.68 | 50.00 | - |
| Discriminative | ChemBERTa and PubMedBERT | S | 42.65 | 25.56 | 31.96 | 87.69 | 60.59 | 79.90 |
| | | SG | 51.07 | 42.00 | 46.09 | 88.88 | 68.43 | 84.92 |
| | | SGE | 51.65 | 55.40 | **53.46** | 89.08 | **74.39** | **88.34** |
| Generative | OpenBioLLM-8B | S | 44.55 | 16.57 | 24.16 | 88.22 | 56.97 | - |
| | | SG | 49.14 | 50.36 | 49.74 | 88.48 | 71.86 | - |
| | | SGE | **52.18** | 54.75 | 53.43 | **89.20** | 74.17 | - |

Supplementary Table 3. Performance of ChemBERTa & PubMedBERT, and OpenBioLLM-8B on the CT-ADE-SOC test set using S, SG, and SGE feature sets. The AUROC metric cannot be computed for baseline and generative models because these models do not produce raw probabilities. MAJ: Majority class prediction; S: SMILES; SG: SMILES and group description; SGE: SMILES, group description, and eligibility criteria. Results are micro-averaged.

| Abbreviation | System organ class (SOC) |
|---|---|
| Blood | Blood and lymphatic system disorders |
| Card | Cardiac disorders |
| Cong | Congenital, familial and genetic disorders |
| Ear | Ear and labyrinth disorders |
| Endo | Endocrine disorders |
| Eye | Eye disorders |
| Gastr | Gastrointestinal disorders |
| Genrl | General disorders and administration site conditions |
| Hepat | Hepatobiliary disorders |
| Immun | Immune system disorders |
| Infec | Infections and infestations |
| Inj&P | Injury, poisoning and procedural complications |
| Inv | Investigations |
| Metab | Metabolism and nutrition disorders |
| Musc | Musculoskeletal and connective tissue disorders |
| Neopl | Neoplasms benign, malignant and unspecified (incl cysts and polyps) |
| Nerv | Nervous system disorders |
| Preg | Pregnancy, puerperium and perinatal conditions |
| Psych | Psychiatric disorders |
| Renal | Renal and urinary disorders |
| Repro | Reproductive system and breast disorders |
| Resp | Respiratory, thoracic and mediastinal disorders |
| Skin | Skin and subcutaneous tissue disorders |
| SocCi | Social circumstances |
| Surg | Surgical and medical procedures |
| Vasc | Vascular disorders |
| Prod | Product issues |

Supplementary Table 4. Mapping of system organ class (SOC) abbreviations to their full terms.

| **Abbreviation** | **ATC full term** |
|---|---|
| A | Alimentary tract and metabolism |
| B | Blood and blood forming organs |
| C | Cardiovascular system |
| D | Dermatologicals |
| G | Genito urinary system and sex hormones |
| H | Systemic hormonal preparations, excl. sex hormones and insulins |
| J | Antiinfectives for systemic use |
| L | Antineoplastic and immunomodulating agents |
| M | Musculo-skeletal system |
| N | Nervous system |
| P | Antiparasitic products, insecticides and repellents |
| R | Respiratory system |
| S | Sensory organs |
| V | Various |

Supplementary Table 5. Mapping of anatomical therapeutic chemical (ATC) main pharmacological group abbreviations to their full terms.

| Label | Precision (%) | Recall (%) | F1-score (%) | Accuracy (%) | Balanced Accuracy (%) | AUROC (%) |
|---|---|---|---|---|---|---|
| Blood and lymphatic system disorders | 56.36 | 44.29 | 49.60 | 90.00 | 70.00 | 87.45 |
| Cardiac disorders | 27.03 | 37.74 | 31.50 | 86.19 | 64.19 | 75.03 |
| Congenital, familial and genetic disorders | 0.00 | 0.00 | 0.00 | 99.76 | 50.00 | 70.38 |
| Ear and labyrinth disorders | 26.67 | 10.00 | 14.55 | 96.27 | 54.55 | 82.77 |
| Endocrine disorders | 37.50 | 13.04 | 19.35 | 98.02 | 56.32 | 70.68 |
| Eye disorders | 43.48 | 27.03 | 33.33 | 90.48 | 61.82 | 80.39 |
| Gastrointestinal disorders | 62.37 | 84.99 | 71.95 | 73.02 | 74.89 | 82.41 |
| General disorders and administration site conditions | 57.86 | 60.07 | 58.94 | 80.87 | 73.55 | 83.43 |
| Hepatobiliary disorders | 11.11 | 3.85 | 5.71 | 97.38 | 51.60 | 78.35 |
| Immune system disorders | 100.00 | 5.56 | 10.53 | 98.65 | 52.78 | 63.66 |
| Infections and infestations | 51.01 | 75.75 | 60.96 | 71.75 | 72.93 | 80.77 |
| Injury, poisoning and procedural complications | 25.93 | 19.81 | 22.46 | 88.49 | 57.31 | 74.28 |
| Investigations | 48.94 | 59.15 | 53.56 | 80.87 | 72.50 | 81.64 |
| Metabolism and nutrition disorders | 34.68 | 54.09 | 42.26 | 81.35 | 69.69 | 81.66 |
| Musculoskeletal and connective tissue disorders | 43.10 | 55.41 | 48.48 | 78.41 | 69.49 | 79.99 |
| Neoplasms benign, malignant and unspecified | 0.00 | 0.00 | 0.00 | 97.62 | 49.96 | 88.40 |
| Nervous system disorders | 64.35 | 79.89 | 71.28 | 73.33 | 74.29 | 82.94 |
| Pregnancy, puerperium and perinatal conditions | 75.00 | 42.86 | 54.55 | 99.60 | 71.39 | 90.71 |
| Product issues | 0.00 | 0.00 | 0.00 | 99.44 | 49.92 | 75.00 |
| Psychiatric disorders | 69.41 | 32.07 | 43.87 | 88.02 | 64.82 | 88.32 |
| Renal and urinary disorders | 40.00 | 14.29 | 21.05 | 91.67 | 56.24 | 77.71 |
| Reproductive system and breast disorders | 44.44 | 8.16 | 13.79 | 96.03 | 53.88 | 74.66 |

| Respiratory, thoracic and mediastinal disorders | 47.09 | 36.32 | 41.01 | 81.51 | 63.77 | 76.32 |
|---|---|---|---|---|---|---|
| Skin and subcutaneous tissue disorders | 42.93 | 43.39 | 43.16 | 82.86 | 66.60 | 77.88 |
| Social circumstances | 0.00 | 0.00 | 0.00 | 99.92 | 50.00 | 66.44 |
| Surgical and medical procedures | 44.44 | 26.67 | 33.33 | 98.73 | 63.13 | 69.81 |
| Vascular disorders | 41.61 | 41.36 | 41.49 | 85.00 | 66.40 | 79.96 |

Supplementary Table 6. Discriminative SGE performance metrics per label on CT-ADE-SOC test set.

| **ATC Category** | **Precision (%)** | **Recall (%)** | **F1-score (%)** | **Accuracy (%)** | **Balanced Accuracy (%)** | **AUROC (%)** |
|---|---|---|---|---|---|---|
| Alimentary Tract and Metabolism | 40.97 | 46.97 | 43.76 | 92.43 | 71.22 | 88.61 |
| Blood and Blood Forming Organs | 45.71 | 69.57 | 55.17 | 88.67 | 80.18 | 86.81 |
| Cardiovascular System | 49.41 | 41.18 | 44.92 | 92.05 | 68.79 | 87.27 |
| Dermatologicals | 47.66 | 43.88 | 45.69 | 92.54 | 70.09 | 88.82 |
| Genito Urinary System and Sex Hormones | 42.67 | 40.51 | 41.56 | 95.67 | 69.18 | 89.81 |
| Systemic Hormonal Preparations | 48.00 | 75.00 | 58.54 | 90.31 | 83.42 | 90.59 |
| Antiinfectives for Systemic Use | 26.77 | 40.96 | 32.38 | 90.44 | 67.17 | 86.59 |
| Antineoplastic and Immunomodulating Agents | 57.35 | 57.88 | 57.62 | 79.85 | 72.27 | 82.76 |
| Musculo-Skeletal System | 41.86 | 44.63 | 43.20 | 90.77 | 69.67 | 87.73 |
| Nervous System | 56.08 | 59.72 | 57.84 | 91.57 | 77.35 | 90.66 |
| Antiparasitic Products, Insecticides and Repellents | 38.46 | 45.98 | 41.88 | 90.86 | 70.15 | 87.46 |
| Respiratory System | 51.95 | 47.62 | 49.69 | 94.12 | 72.38 | 90.73 |
| Sensory Organs | 60.87 | 50.00 | 54.90 | 97.63 | 74.52 | 92.76 |
| Various | 49.47 | 55.95 | 52.51 | 92.32 | 75.63 | 87.47 |
| No ATC | 52.96 | 57.20 | 55.00 | 87.58 | 74.71 | 87.72 |

Supplementary Table 7. Discriminative SGE performance metrics by ATC main pharmacological groups on the CT-ADE-SOC test set.